# Nearly Solved? Robust Deepfake Detection Requires More than Visual Forensics

Guy Levy
Balvatnik School of Computer Science
Tel-Aviv University
Tel-Aviv
gl2@mail.tau.ac.il

Nathan Liebmann
Balvatnik School of Computer Science
Tel-Aviv University
Tel-Aviv
nathanlibman@mail.tau.ac.il

**Abstract**

Deepfakes are on the rise, with increased sophistication and prevalence allowing for high-profile social engineering attacks. Detecting them in the wild is therefore important as ever, giving rise to new approaches breaking benchmark records in this task. In line with previous work, we show that recently developed state-of-the-art detectors are susceptible to classical adversarial attacks, even in a highly-realistic black-box setting, putting their usability in question. We argue that crucial 'robust features' of deepfakes are in their higher semantics, and follow that with evidence that a detector based on a semantic embedding model is less susceptible to black-box perturbation attacks. We show that large visuo-lingual models like GPT-4o can perform zero-shot deepfake detection better than current state-of-the-art methods, and introduce a novel attack based on high-level semantic manipulation. Finally, we argue that hybridising low- and high-level detectors can improve adversarial robustness, based on their complementary strengths and weaknesses.

## 1. Introduction

### 1.1. Deepfake detection

The impact of deepfake is no longer deniable. From inflammatory words put into politicians' mouths [19] to highly targeted phishing attacks on corporations [5, 32, 40], the ability to deceive by providing synthesised evidence is on the rise [41] and the importance of detecting its occurrences and inner workings is as high as ever. As deepfake technology continues to advance, its potential to influence public opinion, manipulate information, and undermine trust in digital communication grows exponentially.

In the realm of deep learning research, adversarial and counter-adversarial techniques evolve in a symbiotic relationship, where each innovation in one direction encourages the development of more advanced approaches in the other.

### 1.2. GAN-based methods

One of the primary methods that proved successful enough to bring the concept of deepfake to public awareness is Generative Adversarial Networks (GANs). This method, introduced by Goodfellow et al. in 2014 [12], consists of two neural networks – a generator and a discriminator – locked in an adversarial training process. The generator creates synthetic images, while the discriminator attempts to distinguish between real and fake images. This adversarial process leads to increasingly realistic synthetic content. In the context of deepfakes, GANs were quickly adapted for face synthesis and manipulation [21].

The evolution of deepfake detection initially centered around approaches specifically designed to identify GAN-generated content. Early detection methods often employed CNNs trained on datasets of real and GAN-generated images. For instance, XceptionNet [10], was adapted for deepfake detection and showed promising results at the time [34]. These initial detectors were typically trained on datasets containing known GAN-generated images, such as those created by popular face-swapping algorithms like FakeApp or DeepFakes.

### 1.3. Diffusion methods and cross-method generalisation

With the introduction of newer kinds of image generators like diffusion models, it was discovered that earlier approaches tacitly focused on artefacts of the GAN based generators used to produce the deepfake content, and therefore did not generalise well to images generated by the new kinds of models. It was then suggested that learning the downstream real/fake classification task should be done on top of an embedding of the images produced by a model *not* trained specifically for classification at all, like the visual encoder of CLIP [26, 30], essentially shifting the focus towards more global and robust features, a key aspect we will discuss below.

A later work built upon this notion and explored several forms of adaptation methods for universal deep fake detection based on CLIP's visual and text encoders [20]. Best results were achieved with the prompt tuning *Context Optimisation (CoOp)* method [43], where learnable vectors are inserted into the textual prompt during downstream task fine-tuning, while the weights of both text and image encoders are frozen.

In a very recent work [7], authors showcased a deepfake detection method which is based on a variant of ResNet50. This variant is trained to identify deepfake instances based on 9x9 patches of an image providing a patch-based deepfake score that is average-pooled to a total deepfake score for an image. This rather simple approach proved to be extremely effective, significantly improving over the state-of-the-art in multiple benchmarks.

Each breakthrough raises the question of inherent superiority - are deep-learning models susceptible to those various adversarial approaches merely because of their design, or can the effectiveness of such techniques be attributed to incomplete training procedure. By some accounts, adversarial examples will always pose an inherent threat to deep-learning algorithms, and specifically deepfake detectors [6]. Yet novel works with benchmark-breaking detection performance propose that deepfake might be a soon-to-be-solved problem in deep-learning [7].

## 2. Related work

### 2.1. Attacking deepfake detectors

Forensic approaches to detect deepfake images using neural network based classifiers have been shown to be susceptible to several types of adversarial attacks, both with direct access to the model (*white-box* setting) and with only query-based access to it (*black-box* setting). Such attacks were shown to reduce the accuracy of state-of-the-art detectors at the time by $\approx$ 70%-99.9% [6].

Based on evidence that deepfake detectors tend to rely on statistical differences between real and fake images, one recent work proposed an evasion mechanism that better generalises to different detectors, by explicitly minimising the statistical difference between the two while generating adversarial examples [14].

This lack of robustness has been shown for deepfake video detectors as well, in the form of successful adversarial attacks imperceptibly modifying already-generated fake content in a way that makes it evade detectors, even under common video compression codecs, resulting in a realistic attack scheme [16].

One of the challenges of real-world deepfake detection is inference performance. Large and complex models might not be a good fit for the availability and speed needed to detect fake artefacts on webpages and social media accessed heavily through end-devices with low computational power. This stresses the importance of distilling smaller and efficient detectors [22].

## 3. Contribution

In the present work we aim to show that recent approaches to deepfake detection [7, 20] are susceptible to well-known adversarial attacks, in both white-box and black-box settings. Specifically, the approach taken by the authors of [7] constrained their model to base its predictions on highly localised features, somewhat analogous to the notion of *non-robust features* [17]. We hypothesise this makes their model highly susceptible to the classical bounded input perturbation attack scheme. Crucially, this poses a serious threat to the usability of such detectors as they are, considering that the context of deepfake detection is adversarial in nature.

On the path towards more robust detection, we show that the approach of [20], which relies on a visuo-lingual general-purpose embedding model, is not as susceptible to simple black-box perturbation attacks.

Next, we pick GPT-4o as a representative of proprietary large visuo-lingual models. We start by showing that a simple zero-shot task with GPT-4o outperforms current state-of-the-art deepfake detection approaches on the highly challenging Celeb-DF dataset. We then introduce a novel typographic attack which aims to confuse the model as for the broader context of the input image. This attack brings down performance on the zero-shot task. Interestingly, even when prompted to specify the justification for the classification verdict, the model does not mention the typographic artefact in many cases where the verdict is flipped by the attack.

Finally, we build the case for hybridising detection methods based on visual artefacts with methods relying on high-level semantics, due to their conflicting vulnerabilities and strengths.

### 3.1. Author Contributions

G.L. led the implementation of model training pipelines and evaluation frameworks for both the LaDeDa and CLIPping the Deception detectors. N.L. developed and implemented the adversarial attack methods. Both authors jointly contributed to dataset curation, experimental design, analysis of results, and manuscript preparation.

## 4. Dataset preparation

For our evaluation, we selected two of the most challenging and widely used datasets in contemporary deepfake detection research. The first dataset is FaceForensics++ [35], which contains 1,000 original videos manipulated using four different methods: Deepfakes [1], Face2Face [39], FaceSwap [2] and NeuralTextures [38]. We uniformly sampled approximately 100,000 frames from these videos, split into train, validation and test sets. Each manipulated video in FaceForensics++ combines two real videos - a source and a target - where facial features from the source (either the full face or just the expression) are transferred to the target video. Since not all possible source-target actor pairs appear in the fake videos, and the appearance frequency of actors in the different sets varies significantly, there is a risk of train-test contamination where such differences in actor appearance frequencies could lead models to learn actor identification rather than deepfake detection. We addressed this concern in our splitting algorithm, by ensuring that if a source-target actor pair appears in a fake video in one of the sets (train/validation/test), all videos containing either actor will be in the same set. Another choice we made is regarding the proportion of real to fake videos in the dataset. The original dataset includes approximately 10 manipulated videos for each source video. For the training procedures we will describe below, we sampled a class-balanced subset of the frames.

The second dataset we chose to use for evaluation is Celeb-DF [24]. This dataset is among the most difficult datasets available to date, and is usually used for evaluation of deepfake detection trained using a different dataset, such as FaceForensics++. The Celeb-DF dataset comprises 590 real videos sampled from YouTube interviews of 59 celebrities, and 5,639 corresponding DeepFake videos generated using an improved DeepFake synthesis algorithm. The improvements include higher resolution face synthesis, enhanced colour consistency through data augmentation and post-processing, and temporal smoothing of facial landmarks. The dataset captures diversity across gender (56.8% male, 43.2% female), age groups (ranging from under 30 to over 60), and ethnicities (88.1% Caucasian, 6.8% African American, 5.1% Asian). The real videos exhibit considerable variation in face size, orientation, lighting conditions, and background settings. The enhanced visual quality of Celeb-DF is confirmed by its higher Mask-SSIM score[1] (0.92) compared to previous datasets (0.80-0.88), making it a particularly challenging benchmark for DeepFake detection methods. For this dataset, we uniformly sampled around 3,000 frames from the official test set videos.

## 5. Black box attack – a genetic approach

For our black-box attack setting, we employ a genetic algorithm (GA) based approach inspired by [3]. Genetic algorithms belong to the broader family of evolutionary algorithms and are particularly suitable for black-box optimisation problems where gradient information is unavailable. In our context, we only require query access to the target model's predictions, making GA an appropriate choice for attacking real-world deepfake detectors. Our implementation follows the standard GA framework with several adaptations for the adversarial attack scenario. The algorithm maintains a population of $n$ candidate adversarial examples, each representing a perturbed version of the input image bounded by $\varepsilon$. During each of the $m$ generations, the algorithm:

1. Evaluates the fitness of each candidate by querying the target model and computing a loss based on the predicted probability of the desired target class.
2. Selects the top-$k$ performing candidates as *elites* that directly survive to the next generation.
3. Creates new candidates through *crossover* operations between *elite* pairs, where successful perturbations from both *parents* are combined.
4. Applies random *mutations* with probability $p$ and weight $w$ to explore the local neighbourhood of successful candidates.

The mutation step is crucial for exploring the attack space while respecting the $\varepsilon$-bound constraint. The algorithm terminates either when it reaches the maximum number of generations or achieves a successful attack. [8] demonstrated that evolutionary approaches can generate adversarial examples that are both effective and visually similar to the original inputs, which is essential in our setting where the attacks should not be visually detectable.

All the attacks we performed were targeted at flipping true-positive predictions (fake frames classified as fake) to false-negatives, which is the realistic scenario for deepfake detection evasion. Consequently, we computed the *attack success rate (ASR)* as the proportion of true-positive the attack managed to flip. One implication of this choice is that overall metrics may not be degraded as much as they would be had we targeted the attack at both classes.

## 6. Attacking LaDeDa – a visual artefact detector

LaDeDa [7] represents a recent forensic approach to deepfake detection that achieved state-of-the-art performance through explicit focus on local image artefacts. The detector segments input images into 9×9 pixel patches, generates per-patch deepfake scores, and pools these scores for image-level classification. This design deliberately constrains the model's receptive field by replacing standard 3×3 convolutions in ResNet50 with 1×1 convolutions, forcing predictions to rely solely on highly localised features.

---
[1] Which essentially measures the similarity of the deepfake's head region to the original video [4, 24].

We hypothesise that this architectural choice, while effective on benchmarks, creates an inherent vulnerability to adversarial attacks. By examining only small patches independently, the model likely becomes sensitive to pixel-level perturbations that can disrupt its local feature detection, even when these perturbations are imperceptible at the image level. This susceptibility aligns with the concept of non-robust features [17] - characteristics that are predictive but easily manipulated through small input modifications.

### 6.1. Training Procedure

For our evaluation, we train LaDeDa on FaceForensics++ with balanced class distribution using the Adam optimiser. We set an initial learning rate of $1 \times 10^{-3}$ and employ an adaptive learning rate schedule that halves the rate when validation metrics plateau, with a minimum learning rate threshold of $1 \times 10^{-6}$. Training is terminated after 5 epochs without improvement in validation performance to prevent overfitting. Under this configuration, the model converged after 75 epochs of training.

### 6.2. Results

Our experiments reveal substantial degradation in LaDeDa's performance on real-world datasets. When evaluated on FF++, the model achieves an accuracy of 69.41%, average precision of 92.65%, and AUC score of 79.63%. Performance deteriorates when evaluating on the more challenging Celeb-DF dataset, where accuracy drops to 62.7% and AUC to not-better-than-chance 48.8%, with an average precision of 65%. These metrics stand in stark contrast to the near-perfect performance reported by the authors on standard benchmarks, and the state-of-the-art results reported on their novel WildRF dataset. Importantly, the latter contains a wide distribution of photorealistic images, drawings, and computer generated graphics, and does not focus on actual deepfakes of humans. We argue that terminology is key here, and a distinction ought to be made between AI-generated content (AIGC) [28] – that in many cases in the WildRF dataset is not photorealistic – and deepfakes, which are realistic images intended to deceive a human to think they are authentic. The significant gap between benchmark and real-world performance suggests that LaDeDa's reliance on local patches makes it particularly vulnerable to high-quality deepfakes where telltale artefacts are less pronounced. The poor generalisation from FaceForensics++ to Celeb-DF suggests that the model learns dataset-specific patterns rather than relying on robust features. Crucially, this focus on highly localised features indicates potential vulnerability to adversarial attacks, as we will demonstrate in the following section.

Initial white-box attack attempts on the WildRF pretrained LaDeDa model using the classical PGD method [25] provided compelling visual evidence of its structural vulnerabilities. The generated adversarial perturbations clearly revealed the model's 9×9 patch boundaries (Figure 1), directly exposing how the detector's rigid segmentation approach can be exploited.

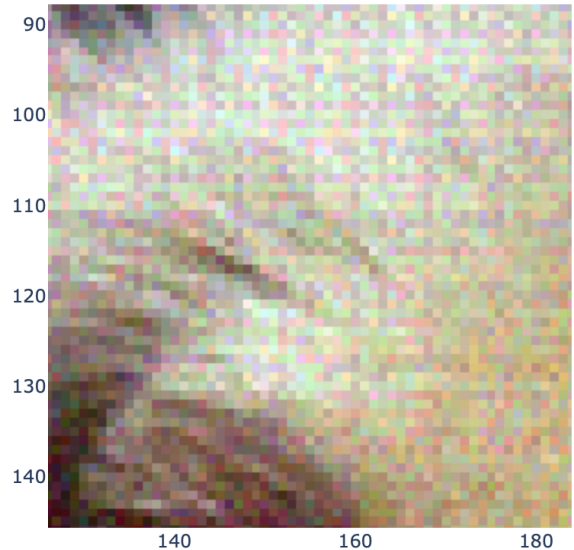

Figure 1: Close-up on image 0006 of the WildRF reddit test set, subject to a PGD attack. A 9×9 grid pattern is visible in the perturbation.

Next, we ran our genetic algorithm black-box attack on the FaceForensics++ test set, with $m = 100, n = 10, k = 5, \varepsilon = 10 \cdot 255^{-1}$. Overall attack success rate was about 70%, performing on average 1,010 queries per input image batch. Note that due to the poor benign performance on Celeb-DF we did not conduct an attack on it.

## 7. Attacking CLIPping the Deception – a visuo-lingual embedding based detector

CLIPping the Deception [20] represents a fundamentally different approach to deepfake detection by leveraging visuo-lingual models. Rather than focusing solely on visual artefacts, this method builds upon CLIP [31], a model trained to align visual and textual representations through contrastive learning on 400 million image-text pairs. This choice is motivated by evidence that visuo-lingual models develop robust and generalisable visual representations through their exposure to diverse, naturally occurring image-text relationships [42]. The use of visuo-lingual models for deepfake detection is particularly appealing as these models have demonstrated strong zero-shot transfer capabilities across various visual tasks without specific training for forensic analysis [37]. Their representations capture high-level semantic concepts, potentially offering a more comprehensive basis for detecting synthetic content.

## 7.1. Training Procedure

Following [20], we employ Context Optimisation (CoOp), a prompt tuning strategy that introduces learnable context tokens while keeping the base CLIP model frozen. This approach offers several advantages. The first of which is efficiency. During training only the context tokens (approximately 12K parameters) require optimisation, compared to the millions of parameters in the CLIP model. Additionally, this method facilitates the preservation of CLIP's robust features – by keeping the model frozen, we maintain the rich visual representations learned during pre-training. Lastly, unlike pure visual approaches, prompt tuning allows the model to leverage CLIP's text encoder for classification. The method perpends a learnable context vector to the class tokens ("real" and "fake"), optimising this vector during training while both visual and text encoders remain frozen. We used a single context vector for both classes and 16 context tokens as recommended in the original work.

We employed this training regiment on the same balanced FaceForensics++ training set that we used to re-train LaDeDa.

## 7.2. Results

Our evaluation on real-world datasets reveals both strengths and limitations of this approach. On the FaceForensics++ test set, the model achieves an accuracy of 60.5%, average precision of 85.52%, and AUC score of 63.82%. On the more challenging Celeb-DF dataset, performance is significantly degraded, but not as catastrophically as with LaDeDa, with accuracy at 56.14%, average precision at 73.92%, and AUC at 58.84%.

The better generalisation of this approach suggests visuolingual representations may provide some inherent robustness. However, the still-significant performance drop indicates that even these more sophisticated representations face challenges with high-quality deepfakes.

In the black-box attack scenario, run with the same parameters as with LaDeDa, attack success rate (ASR) was dramatically lower - around 30%, 40 points lower than the identical attack on LaDeDa. This remarkable result provides strong evidence that this model is more adversarially robust. We argue that this robustness is linked to its reliance on a high-level semantic embedding space. Full results are presented in Table 1.

## 8. Zero-shot deepfake detection with GPT-4o

To fully examine our hypothesis on the importance of high-level semantics, we devised a zero-shot classification task for GPT-4o, a leading commercial visuo-lingual model [23, 36]. Such models are trained on vast, rich datasets from across the web [15], which endows them with extensive world knowledge [29]. One previous attempt to perform deepfake detection with large VLMs including GPT-4o showed performance on-par and even surpassing leading methods [18]. Importantly, the advantage of their approach was most evident when post-processing (e.g., JPEG compression, Gaussian blur) was applied to input images, implying that VLMs are more robust to low-level visual alterations than traditional methods. The task constructed there prompted the model to analyse the input image for visual artefacts. Evaluation was conducted on datasets containing authentic as well as AI-generated human faces.

Our setup differs from [18] in two important aspects: our task formulation does not instruct the model on how to perform the classification, and in particular we do not ask it to find visual artefacts; and the dataset we evaluated on, Celeb-DF, consists of frames from deepfake videos employing generation tactics common in real-world targeted campaigns, such as face-swapping.

## 8.1. The zero-shot task

We composed the following, basic prompt to go with an input image:

"Is this image real or fake?"

We additionally use the native Structured Outputs feature [27] to constrain the output format to the following structure:

```
{ verdict: "real" | "fake", reason: string }
```

Each context (prompt + input image) is used to generate five independent outputs, and the classification score for the *fake* label is the ratio of *fake* verdicts. Interestingly, in some cases the model refused to respond to our prompt. We speculate this results from guardrails put in place against using the model for face recognition and similar potentially privacy-violating tasks. We handle this issue by simply retrying the request.

## 8.2. Evaluation results

We evaluated our scheme on ⅓ of the Celeb-DF test set[2], sampled according to its native class weights to yield 570 fake and 300 authentic frames. As the results in Table 1 show, this very simple zero-shot task performs drastically better than the two dedicated models we discussed earlier. We hypothesise part of the success could be attributed to the fact that images of some of the celebrities in the dataset appeared in the training data used for the visual encoder of GPT-4o. This potentially allows for complex, semantic mismatch detection, like a known figure appearing in a highly improbable context. Such semantic features are invariant to the deepfake generation method and quality, and could therefore contribute to more robust detection.

---

[2]Evaluation was not performed on the full test set for financial efficiency.

## 9. Typographic image context attack

Large visuo-lingual models are susceptible to a family of attacks dubbed *typographic attacks*, where text in an input image – either added in post-processing or part of photographed scene – could distract the model from visual features, and lead to unexpected and incorrect results on downstream tasks like classification [9, 11].

Inspired by these findings, we propose a kind of typographic attack aimed at making a large VLM misclassify a deepfake in a zero-shot task. For this attack, we add overlay text to the input image that appears like the source path of the image file. The path components are constructed such that they suggest the image is part of a control set of a deepfake detection dataset, suggesting it is an authentic image (Figure 2).

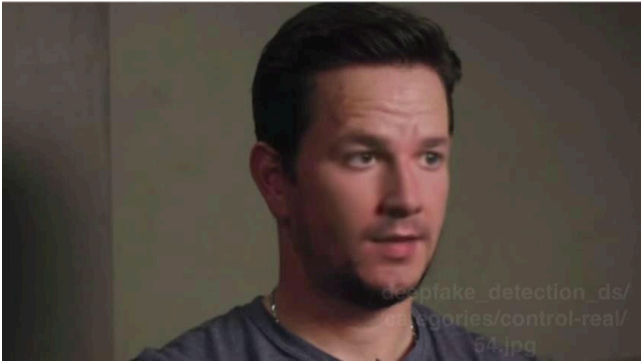

Figure 2: Sample image from the Celeb-DF test set with our attack

After a small-scale pilot, and with the goal of minimising the prominence of the attack to human observers, we settled on white, 7% opacity text of size 29pt at the bottom-right corner of the image. With the attack added to every image of the Celeb-DF evaluation set, we re-evaluated our zero-shot task with GPT-4o on the attacked images. Attack success rate was 6.64%. Interestingly, we noticed that in some cases where the attack succeeded, the reason provided by the model for its classification verdict consistently cited a lack of evidence of manipulation, and did not at all mention the overlayed text that in fact flipped the verdict, a finding that is in-line with some of the results of [9].

While the success rate of our attack was relatively low, we believe it serves as an important proof-of-concept for a black-box attack specifically targeting deepfake detectors relying on multi-modal, semantic embedding, without being obvious to a human observer.

## 10. Discussion

### 10.1. Hybridising visual forensics and high level semantics

As our experiment with GPT-4o shows, large VLMs show great promise as deepfake detectors. However, their reliance on high-level semantics might come at the cost of an inferior ability to detect visual artefacts of deepfake generation; experimental results with detail-oriented visual tasks show that without explicit training-time treatment, VLMs' semantic representation of images omits low-level details [13]. Additionally, comparing CLIP and CNN-based models for AIGC detection on the low-level has shown that the two rely on different parts of the frequency spectrum, and attack transferability between the two is very poor [33]. This is where we believe high-level and low-level detectors present complementary strengths and weaknesses, considering that as discussed the latter lack the ability to attend to semantic, high-level features that are telling of image authenticity. Moreover, adversarial attacks targeting one method might make a fake image *easier* to detect by the other method: overlayed text that confuses a large VLM is easily detectable by a low-level forensics model, for which it is nothing more than a simple disturbance in the image.

| Model | Dataset | Scenario | ACC | Precision | Recall | mAP | ROC AUC | ASR[3] | NQ[4] |
|---|---|---|---|---|---|---|---|---|---|
| LaDeDa | FF++ | Benign | 0.6941 | 0.9118 | 0.6684 | 0.9265 | 0.7963 | - | - |
| | | Genetic BB Attack | 0.3327 | 0.7559 | 0.2 | 0.7189 | 0.2213 | 0.7 | 1,010 |
| | Celeb-DF | Benign | 0.627 | 0.652 | 0.925 | 0.65 | 0.488 | - | - |
| CLIPping the Deception | FF++ | Benign | 0.605 | 0.8252 | 0.6197 | 0.8594 | 0.6382 | - | - |
| | | Genetic BB Attack | 0.4602 | 0.7669 | 0.4321 | 0.8143 | 0.558 | 0.303 | 1,010 |
| | Celeb-DF | Benign | 0.5614 | 0.7095 | 0.5618 | 0.7392 | 0.5884 | - | - |
| GPT-4o zero-shot | Celeb-DF | Benign | 0.6425 | 0.8776 | 0.528 | 0.819 | 0.7318 | - | - |
| | | Typographic Attack | 0.6149 | 0.8593 | 0.493 | 0.7984 | 0.721 | 0.0664 | 0 |

Table 1: Evaluation results

---

[3]Attack success rate, calculated as the number of true-positive flipped to false-negatives

[4]Average number of queries made to the model during the attack, per input

Future work should focus on constructing a hybrid ensemble of low- and high-level models, and explore architectures such as training a linear model to combine the results of the two methods, or adding an adapter model that connects directly to latent representations in both underlying models.

**10.2. Conclusion**

This work demonstrates that despite recent advances in deepfake detection, current approaches remain vulnerable to various forms of adversarial attacks. Our systematic evaluation reveals that local feature-based detectors are particularly susceptible to classical adversarial perturbations. In contrast, models leveraging high-level semantic understanding show greater robustness but might lack the ability to attend to low-level forensic artefacts.

The opposing vulnerabilities of visual forensics and semantic analysis approaches suggest that robust deepfake detection likely requires a more holistic approach combining both. Our findings with GPT-4o's zero-shot capabilities point toward the potential of large visuo-lingual models in this domain, while also highlighting their potentially susceptibility to a new class of attacks, exemplified in our novel typographic image context attack.

Rather than viewing deepfake detection as a soon-to-be-solved problem, we argue that future research should focus on developing hybrid approaches that combine the complementary strengths of different detection paradigms. Such efforts will be crucial as deepfake technology continues to evolve and pose increasingly sophisticated challenges to digital media authenticity.

# References

[1] DeepFakes Github. Retrieved from https://github.com/deepfakes/faceswap

[2] FaceSwap Github. Retrieved from https://github.com/MarekKowalski/FaceSwap/

[3] Moustafa Alzantot, Yash Sharma, Supriyo Chakraborty, Huan Zhang, Cho-Jui Hsieh, and Mani Srivastava. 2019. GenAttack: Practical Black-box Attacks with Gradient-Free Optimization. Retrieved from https://arxiv.org/abs/1805.11090

[4] Illya Bakurov, Marco Buzzelli, Raimondo Schettini, Mauro Castelli, and Leonardo Vanneschi. 2022. Structural similarity index (SSIM) revisited: A data-driven approach. *Expert Systems with Applications* 189, (2022), 116087. https://doi.org/https://doi.org/10.1016/j.eswa.2021.116087

[5] Dylan Butts. 2024. Deepfake scams have robbed companies of millions. Experts warn it could get worse. Retrieved 10 August 2024 from https://www.cnbc.com/2024/05/28/deepfake-scams-have-looted-millions-experts-warn-it-could-get-worse.html

[6] Nicholas Carlini and Hany Farid. 2020. Evading Deepfake-Image Detectors with White- and Black-Box Attacks. Retrieved from https://arxiv.org/abs/2004.00622

[7] Bar Cavia, Eliahu Horwitz, Tal Reiss, and Yedid Hoshen. 2024. Real-Time Deepfake Detection in the Real-World. Retrieved from https://arxiv.org/abs/2406.09398

[8] Jinyin Chen, Mengmeng Su, Shijing Shen, Hui Xiong, and Haibin Zheng. 2019. POBA-GA: Perturbation optimized black-box adversarial attacks via genetic algorithm. *Computers & Security* 85, (August 2019), 89–106. https://doi.org/10.1016/j.cose.2019.04.014

[9] Hao Cheng, Erjia Xiao, Jindong Gu, Le Yang, Jinhao Duan, Jize Zhang, Jiahang Cao, Kaidi Xu, and Renjing Xu. 2024. Unveiling Typographic Deceptions: Insights of the Typographic Vulnerability in Large Vision-Language Model. Retrieved from https://arxiv.org/abs/2402.19150

[10] François Chollet. 2017. Xception: Deep Learning with Depthwise Separable Convolutions. Retrieved from https://arxiv.org/abs/1610.02357

[11] Gabriel Goh, Nick Cammarata †, Chelsea Voss †, Shan Carter, Michael Petrov, Ludwig Schubert, Alec Radford, and Chris Olah. 2021. Multimodal Neurons in Artificial Neural Networks. *Distill* (2021). https://doi.org/10.23915/distill.00030

[12] Ian J. Goodfellow, Jean Pouget-Abadie, Mehdi Mirza, Bing Xu, David Warde-Farley, Sherjil Ozair, Aaron Courville, and Yoshua Bengio. 2014. Generative Adversarial Networks. Retrieved from https://arxiv.org/abs/1406.2661

[13] Chenhui Gou, Abdulwahab Felemban, Faizan Farooq Khan, Deyao Zhu, Jianfei Cai, Hamid Rezatofighi, and Mohamed Elhoseiny. 2024. How Well Can Vision Language Models See Image Details?. Retrieved from https://arxiv.org/abs/2408.03940

[14] Yang Hou, Qing Guo, Yihao Huang, Xiaofei Xie, Lei Ma, and Jianjun Zhao. 2023. Evading DeepFake Detectors via Adversarial Statistical Consistency. In *2023 IEEE/CVF Conference on Computer Vision and Pattern Recognition (CVPR)*, 2023. 12271–12280. https://doi.org/10.1109/CVPR52729.2023.01181

[15] Aaron Hurst, Adam Lerer, Adam P Goucher, Adam Perelman, Aditya Ramesh, Aidan Clark, AJ Ostrow, Akila Welihinda, Alan Hayes, Alec Radford, and others. 2024. Gpt-4o system card. *arXiv preprint arXiv:2410.21276* (2024).

[16] Shehzeen Hussain, Paarth Neekhara, Malhar Jere, Farinaz Koushanfar, and Julian McAuley. 2021. Adversarial deepfakes: Evaluating vulnerability of deepfake detectors to adversarial examples. In *Proceedings of the IEEE/CVF winter conference on applications of computer vision*, 2021. 3348–3357.

[17] Andrew Ilyas, Shibani Santurkar, Dimitris Tsipras, Logan Engstrom, Brandon Tran, and Aleksander Madry. 2019. Adversarial examples are not bugs, they are features. *Advances in neural information processing systems* 32, (2019).

[18] Shan Jia, Reilin Lyu, Kangran Zhao, Yize Chen, Zhiyuan Yan, Yan Ju, Chuanbo Hu, Xin Li, Baoyuan Wu, and Siwei Lyu. 2024. Can chatgpt detect deepfakes? a study of using multimodal large language models for media forensics. In *Proceedings of the IEEE/CVF Conference on Computer Vision and Pattern Recognition*, 2024. 4324–4333.

[19] Huo Jingnan. 2024. It's quick and easy to clone famous politicians' voices, despite safeguards. Retrieved from https://www.npr.org/2024/05/30/nx-s1-4986088/deepfake-audio-elections-politics-ai

[20] Sohail Ahmed Khan and Duc-Tien Dang-Nguyen. 2024. CLIPping the Deception: Adapting Vision-Language Models for Universal Deep-

fake Detection. In *Proceedings of the 2024 International Conference on Multimedia Retrieval*, 2024. 1006–1015.

[21] Pavel Korshunov and Sebastien Marcel. 2018. DeepFakes: a New Threat to Face Recognition? Assessment and Detection. Retrieved from https://arxiv.org/abs/1812.08685

[22] Romeo Lanzino, Federico Fontana, Anxhelo Diko, Marco Raoul Marini, and Luigi Cinque. 2024. Faster Than Lies: Real-time Deepfake Detection using Binary Neural Networks. In *Proceedings of the IEEE/CVF Conference on Computer Vision and Pattern Recognition (CVPR) Workshops*, June 2024. 3771–3780.

[23] Tony Lee, Haoqin Tu, Chi Heem Wong, Wenhao Zheng, Yiyang Zhou, Yifan Mai, Josselin Somerville Roberts, Michihiro Yasunaga, Huaxiu Yao, Cihang Xie, and Percy Liang. 2024. VHELM: A Holistic Evaluation of Vision Language Models. Retrieved from https://arxiv.org/abs/2410.07112

[24] Yuezun Li, Xin Yang, Pu Sun, Honggang Qi, and Siwei Lyu. 2020. Celeb-DF: A Large-scale Challenging Dataset for DeepFake Forensics. Retrieved from https://arxiv.org/abs/1909.12962

[25] Aleksander Madry, Aleksandar Makelov, Ludwig Schmidt, Dimitris Tsipras, and Adrian Vladu. 2019. Towards Deep Learning Models Resistant to Adversarial Attacks. Retrieved from https://arxiv.org/abs/1706.06083

[26] Utkarsh Ojha, Yuheng Li, and Yong Jae Lee. 2023. Towards universal fake image detectors that generalize across generative models. In *Proceedings of the IEEE/CVF Conference on Computer Vision and Pattern Recognition*, 2023. 24480–24489.

[27] OpenAI. 2024. Introducing Structured Outputs in the API. Retrieved from https://openai.com/index/introducing-structured-outputs-in-the-api/

[28] Gan Pei, Jiangning Zhang, Menghan Hu, Zhenyu Zhang, Chengjie Wang, Yunsheng Wu, Guangtao Zhai, Jian Yang, Chunhua Shen, and Dacheng Tao. 2024. Deepfake Generation and Detection: A Benchmark and Survey. Retrieved from https://arxiv.org/abs/2403.17881

[29] Fabio Petroni, Tim Rocktäschel, Sebastian Riedel, Patrick Lewis, Anton Bakhtin, Yuxiang Wu, and Alexander Miller. 2019. Language Models as Knowledge Bases?. In *Proceedings of the 2019 Conference on Empirical Methods in Natural Language Processing and the 9th International Joint Conference on Natural Language Processing (EMNLP-IJCNLP)*, November 2019. Association for Computational Linguistics, Hong Kong, China, 2463–2473. https://doi.org/10.18653/v1/D19-1250

[30] Alec Radford, Jong Wook Kim, Chris Hallacy, Aditya Ramesh, Gabriel Goh, Sandhini Agarwal, Girish Sastry, Amanda Askell, Pamela Mishkin, Jack Clark, and others. 2021. Learning transferable visual models from natural language supervision. In *International conference on machine learning*, 2021. 8748–8763.

[31] Alec Radford, Jong Wook Kim, Chris Hallacy, Aditya Ramesh, Gabriel Goh, Sandhini Agarwal, Girish Sastry, Amanda Askell, Pamela Mishkin, Jack Clark, Gretchen Krueger, and Ilya Sutskever. 2021. Learning Transferable Visual Models From Natural Language Supervision. Retrieved from https://arxiv.org/abs/2103.00020

[32] Nick Robins-Early. 2024. CEO of world's biggest ad firm targeted by deepfake scam. Retrieved 10 August 2024 from https://www.theguardian.com/technology/article/2024/may/10/ceo-wpp-deepfake-scam

[33] Vincenzo De Rosa, Fabrizio Guillaro, Giovanni Poggi, Davide Cozzolino, and Luisa Verdoliva. 2024. Exploring the Adversarial Robustness of CLIP for AI-generated Image Detection. Retrieved from https://arxiv.org/abs/2407.19553

[34] Andreas Rössler, Davide Cozzolino, Luisa Verdoliva, Christian Riess, Justus Thies, and Matthias Nießner. 2018. FaceForensics: A Large-scale Video Dataset for Forgery Detection in Human Faces. Retrieved from https://arxiv.org/abs/1803.09179

[35] Andreas Rössler, Davide Cozzolino, Luisa Verdoliva, Christian Riess, Justus Thies, and Matthias Nießner. 2019. FaceForensics++: Learning to Detect Manipulated Facial Images. In *International Conference on Computer Vision (ICCV)*, 2019.

[36] Sakib Shahriar, Brady Lund, Nishith Reddy Mannuru, Muhammad Arbab Arshad, Kadhim Hayawi, Ravi Varma Kumar Bevara, Aashrith Mannuru, and Laiba Batool. 2024. Putting GPT-4o to the Sword: A Comprehensive Evaluation of Language, Vision, Speech, and Multi-modal Proficiency. Retrieved from https://arxiv.org/abs/2407.09519

[37] Haoyu Song, Li Dong, Wei-Nan Zhang, Ting Liu, and Furu Wei. 2022. CLIP Models are Few-shot Learners: Empirical Studies on VQA and Visual Entailment. Retrieved from https://arxiv.org/abs/2203.07190

[38] Justus Thies, Michael Zollhöfer, and Matthias Nießner. 2019. Deferred Neural Rendering: Image Synthesis using Neural Textures. Retrieved from https://arxiv.org/abs/1904.12356

[39] Justus Thies, Michael Zollhöfer, Marc Stamminger, Christian Theobalt, and Matthias Nießner. 2020. Face2Face: Real-time Face Capture and Reenactment of RGB Videos. Retrieved from https://arxiv.org/abs/2007.14808

[40] Stuart A. Thomson. 2024. How 'Deepfake Elon Musk' Became the Internet's Biggest Scammer. Retrieved 17 August 2024 from https://www.nytimes.com/interactive/2024/08/14/technology/elon-musk-ai-deepfake-scam.html

[41] Mika Westerlund. 2019. The emergence of deepfake technology: A review. *Technology innovation management review* 9, 11 (2019).

[42] Yunqing Zhao, Tianyu Pang, Chao Du, Xiao Yang, Chongxuan Li, Ngai-Man Man Cheung, and Min Lin. 2024. On evaluating adversarial robustness of large vision-language models. *Advances in Neural Information Processing Systems* 36, (2024).

[43] Kaiyang Zhou, Jingkang Yang, Chen Change Loy, and Ziwei Liu. 2022. Learning to prompt for vision-language models. *International Journal of Computer Vision* 130, 9 (2022), 2337–2348.